\documentclass[lettersize,journal]{IEEEtran}
\usepackage{amsmath,amsfonts}
\usepackage{mathrsfs}
\usepackage{array}
\usepackage[caption=false,font=normalsize,labelfont=sf,textfont=sf]{subfig}
\usepackage{textcomp}
\usepackage{stfloats}
\usepackage{url}
\usepackage{graphicx}
\usepackage{cite}
\usepackage{bm}
\usepackage{nomencl}
\usepackage{booktabs}
\usepackage{mathrsfs}
\usepackage{multirow}
\usepackage{makecell}
\usepackage{hyperref}
\usepackage[linesnumbered,ruled,vlined]{algorithm2e}
\usepackage{flushend}
\usepackage{threeparttable}
\usepackage{xcolor}
\usepackage{tabularx}
\usepackage{verbatim}
\usepackage{colortbl}
\def\BibTeX{{\rm B\kern-.05em{\sc i\kern-.025em b}\kern-.08em
    T\kern-.1667em\lower.7ex\hbox{E}\kern-.125emX}}
\usepackage{marvosym}

\hypersetup{hidelinks} 
\usepackage{pifont} 

\begin{document}

\title{Uncertainty-Aware Prediction and Application in Planning for Autonomous Driving: Definitions, Methods, and Comparison}


\author{
Wenbo Shao, Jiahui Xu, Zhong Cao,~\IEEEmembership{Member,~IEEE}, Hong Wang\textsuperscript{\Letter},~\IEEEmembership{Senior Member,~IEEE}, Jun Li
            \thanks{This work has been submitted to the IEEE for possible publication. Copyright may be transferred without notice.
            This work was supported by the National Natural Science Foundation of China Project: 52072215, 52221005 and U1964203, National key R\&D Program of China: 2022YFB2503003, and State Key Laboratory of Intelligent Green Vehicle and Mobility. \textit{(Corresponding Authors: Hong Wang)}}
            \thanks{Wenbo Shao, Zhong Cao, Hong Wang and Jun Li are with School of Vehicle and Mobility, Tsinghua University, Beijing 100084, China. (e-mail: swb19@mails.tsinghua.edu.cn, caozhong@tsinghua.edu.cn, hong\_wang@tsinghua.edu.cn, lijun1958@tsinghua.edu.cn)}
            \thanks{Jiahui Xu is with the School of Mechanical Engineering, Beijing Institute of Technology, Beijing 100081, China. (e-mail: 3120220416@bit.edu.cn).}
}


\maketitle

\color{black}
\begin{abstract}
Autonomous driving systems face the formidable challenge of navigating intricate and dynamic environments with uncertainty. This study presents a unified prediction and planning framework that concurrently models short-term aleatoric uncertainty (SAU), long-term aleatoric uncertainty (LAU), and epistemic uncertainty (EU) to predict and establish a robust foundation for planning in dynamic contexts. 
The framework uses Gaussian mixture models and deep ensemble methods, to concurrently capture and assess SAU, LAU, and EU, where traditional methods do not integrate these uncertainties simultaneously.
Additionally, uncertainty-aware planning is introduced, considering various uncertainties. The study's contributions include comparisons of uncertainty estimations, risk modeling, and planning methods in comparison to existing approaches.
The proposed methods were rigorously evaluated using the CommonRoad benchmark and settings with limited perception. These experiments illuminated the advantages and roles of different uncertainty factors in autonomous driving processes.
In addition, comparative assessments of various uncertainty modeling strategies underscore the benefits of modeling multiple types of uncertainties, thus enhancing planning accuracy and reliability. The proposed framework facilitates the development of methods for UAP and surpasses existing uncertainty-aware risk models, particularly when considering diverse traffic scenarios.
Project page: \href{https://swb19.github.io/UAP/}{https://swb19.github.io/UAP/}.

\end{abstract}


\color{black}
\section{Introduction}

\begin{figure}
    \centering
    \includegraphics[width=9cm]{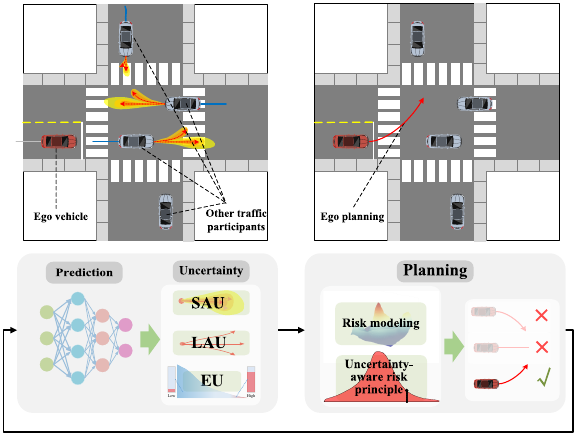}
    \caption{Uncertainty in prediction and risk-aware planning.}
    \label{Fig1}
\end{figure}

\IEEEPARstart{T}{he} rapid evolution of autonomous driving technology has significantly impacted the automotive industry. The key to autonomous driving is the ability to predict nearby traffic motions and thereby make safe and efficient decisions. The prediction module, fed by inputs from perception and localization systems, is crucial for forecasting the movements of objects around an autonomous vehicle and provides vital data for navigating complex scenarios \cite{huang2022survey}. Recent advancements in data-driven technologies, particularly deep neural networks (DNNs), have significantly improved the prediction and planning capabilities.

However, accurately predicting movements and associated risks remains challenging, primarily due to uncertainty \cite{li2023key}. In artificial intelligence (AI), uncertainty is divided into aleatoric uncertainty (AU) and epistemic uncertainty (EU) \cite{kendall2017uncertainties}. For predictive tasks, uncertainty is categorized into:
1) Short-term AU (SAU), arising from the stochastic nature of traffic participant motion and perceptual noise, representing a short-term and concentrated form of uncertainty. For instance, because of disturbances from driver-operated signals or environmental factors, a vehicle's control command at a given moment may exhibit a limited degree of randomness. Additionally, noise introduced during data acquisition or annotation processes may also contribute to SAU. 
2) Long-term AU (LAU), resulting from the diversified behavioural patterns of traffic participants, presents a multimodal form of uncertainty. For example, at an intersection, a vehicle may choose from various actions such as turning left, proceeding straight, or employing different route strategies for a left turn. 
3) EU, stemming from inadequate training, manifests as an inherent uncertainty of the model that relies on training data. For instance, the prediction model may exhibit cognitive bias when faced with rare or absent scenarios in the training data, leading to high EU. These uncertainties directly affect the prediction results, subsequently influencing the effectiveness of risk quantification and planning.

Previous research has often focused on single uncertainty types using approaches such as unimodal Gaussian distributions for SAU and multimodal predictions for LAU \cite{cheng2023gatraj, chen2022scept}, with increasing attention on EU extraction. However, integrated modeling of these uncertainties has been less explored. In motion planning, methods addressing AU or EU exist, but there is a lack of comprehensive frameworks that integrate these uncertainties \cite{zhou2022long} and systematic comparative discussions. Moreover, it is difficult to discern the strengths and limitations of various uncertainties; therefore, a need for a comprehensive framework that addresses all forms of uncertainty in prediction-planning systems is required.
This study aimed to fill this gap by proposing a holistic approach, as illustrated in Fig. \ref{Fig1}, introducing methods that model SAU, LAU, and EU concurrently, establish an uncertainty-sensitive planning framework, and compare various uncertainty modeling and risk quantification techniques.
The primary contributions are:

1) \textbf{Definition and modeling of three types of uncertainty in predictions}: A novel integrated network architecture is proposed to simultaneously model all or parts of the SAU, LAU, and EU in the prediction. This enhances prediction reliability and provides richer information and a more robust foundation for planning in a dynamic environment.

2) \textbf{Uncertainty-aware planning (UAP)}: A prediction-planning framework that considers various uncertainties is proposed. Targeted risk models are designed for different uncertainties, allowing the planning module to consider multiple uncertainty factors. This ensures that the prediction-planning system operates effectively to address dynamic and uncertain real-world driving scenarios and the inherent potential functionality limitations of the prediction module.

3) \textbf{Comprehensive comparisons for analyzing various uncertainties and their roles in planning}: Based on the proposed prediction-planning framework, comprehensive experiments are designed to compare the advantages of different uncertainties, risk models, and UAP methods. This provides a clear and systematic perspective and inspiration.

The remainder of this paper is structured as follows: Section II reviews related work. Section III defines the prediction-planning problem. Section IV introduces the proposed prediction model with multiple types of uncertainty estimates, followed by the UAP method in Section V. The experimental setup and analysis are presented in Sections VI and VII, respectively. Section VIII concludes this paper.

\section{Related Work}
\subsection{Prediction for Autonomous Driving}
Prediction is essential for autonomous driving, enabling proactive strategy formulation during planning. Early research focused on simple kinematic models such as constant velocity or Kalman filters, which offer interpretability but are limited in complex scenarios \cite{lefevre2014survey}. Machine learning methods later improved medium-to long-term predictions yet struggled with highly dynamic environments \cite{huang2022survey}.

The rise of DNNs, particularly recurrent neural networks (RNNs), long short-term memory (LSTM), and gated recurrent units (GRUs), has revolutionised trajectory prediction. These models effectively capture temporal dependencies in complex traffic conditions \cite{huang2022survey, 10186629}. Transformers utilize self-attention mechanisms and further enhance the modeling of temporal features \cite{zhou2022hivt}.
Graph-based models and convolutional social pooling have emerged to better understand spatial interactions, significantly improving the accuracy in urban settings \cite{sheng2022graph, geisslinger2021watch}. Incorporating map information, such as vector maps and bird's-eye views, has also been vital for a comprehensive understanding of various scenarios, aiding more informed predictions \cite{gao2020vectornet,chou2020predicting}. Building on existing research, this study employed a representative trajectory prediction model designed to simultaneously consider information from traffic participants and a map.

\subsection{Modeling of Prediction Uncertainty}
Prediction uncertainty in autonomous driving, encompassing SAU, LAU, and EU, significantly influences trajectory prediction and planning.

SAU, characterised by its transient nature and relatively centralised distribution, is typically modeled using multivariate unimodal Gaussian distributions to enhance the robustness of trajectory predictions \cite{ma2019trafficpredict}. LAU, which reflects a multimodal distribution of potential trajectories owing to varied traffic behaviours \cite{huang2023multimodal}, is addressed using diverse methods. Generative models, such as GANs \cite{lv2022improved} and CVAEs \cite{ivanovic2020multimodal} effectively capture behaviour variability. Techniques such as TNT \cite{zhao2021tnt} and denseTNT \cite{gu2021densetnt} use target-guided strategies for LAU modeling. Gaussian mixture models (GMMs) \cite{zheng2021unlimited} offer a direct and effective approach to represent this uncertainty, capturing multiple outcomes over an extended timeframe.

Data-driven prediction models that rely on training data often struggle with rare or unseen scenarios and spatiotemporal shifts, thereby leading to potential performance inadequacies \cite{gawlikowski2023survey,10143381}. EU reflects the model's comprehension of input data and addresses these limitations. Various methods for EU estimation have evolved, from early Gaussian processes and Markov chain Monte Carlo \cite{abdar2021review} to modern DNN-based techniques like Monte Carlo dropout \cite{gal2016dropout}, deep evidential regression \cite{amini2020deep}, and deep ensemble \cite{lakshminarayanan2017simple, cao2023continuous}. In particular, deep ensemble and its variants \cite{wen2020batchensemble} have proven effective in assessing EU.

Few existing studies have concurrently modeled and compared the three types of uncertainties in the prediction models. Focusing solely on one or two uncertainties may lead to an incomplete understanding of predictive uncertainty. This study aims to collectively model and evaluate these three uncertainties, offering valuable insights into the influence of various uncertainties and their combinations in improving the reliability of autonomous driving systems.

\subsection{Uncertainty-Aware Planning}
In autonomous driving, planning under uncertainty is a pivotal aspect that profoundly influences the system safety and effectiveness. This involves addressing various types of uncertainty, including SAU, LAU, and EU, to ascertain optimal decisions.

Dealing with SAU in planning has garnered attention. One approach involves integrating probabilistic forecasts \cite{geisslinger2023ethical,xu2023risk,khaitan2021safe,hardy2010contingency,xu2014motion}, which accounts for the inherent unpredictability of immediate future trajectories. 

LAU presents a substantial challenge owing to its variety of potential future maneuvers. Addressing LAU effectively necessitates the ability to evaluate multiple futures over an extended timeframe \cite{cui2021lookout}. Furthermore, certain studies have not clearly distinguished between SAU and LAU in their modeling, instead opting to model behavioural uncertainties directly and implement strategic planning \cite{ding2021epsilon,li2023marc,10161419,hubmann2018automated}.
In this study, the implementation of GMM strategies is recommended for the explicit modeling of both SAU and LAU. Here, the study concurrently refines the predicted outcomes and weights for all modes during the training process, while providing detailed insights into the predicted results and likelihood of different modes within the planning framework.

In recent years, the role of EU in mitigating potential model shortcomings or long-tail scenarios has been explored and discussed \cite{hoel2023ensemble,yang2023towards,tang2022prediction}. Metrics such as failure risk and tail exponent have been introduced to quantify EU risks. There are two primary approaches for handling this uncertainty in planning. One involves modeling it as a multivariate unimodal Gaussian distribution, subsequently establishing safety constraints or modeling safety risks \cite{10184920}. Another approach leverages methods such as conditional value at risk (CVaR) to achieve risk-sensitive planning \cite{9971752}. However, there is a tendency to overlook the combined impact of SAU and LAU in current EU management strategies.

\begin{figure}
    \centering
    \includegraphics[width=5cm]{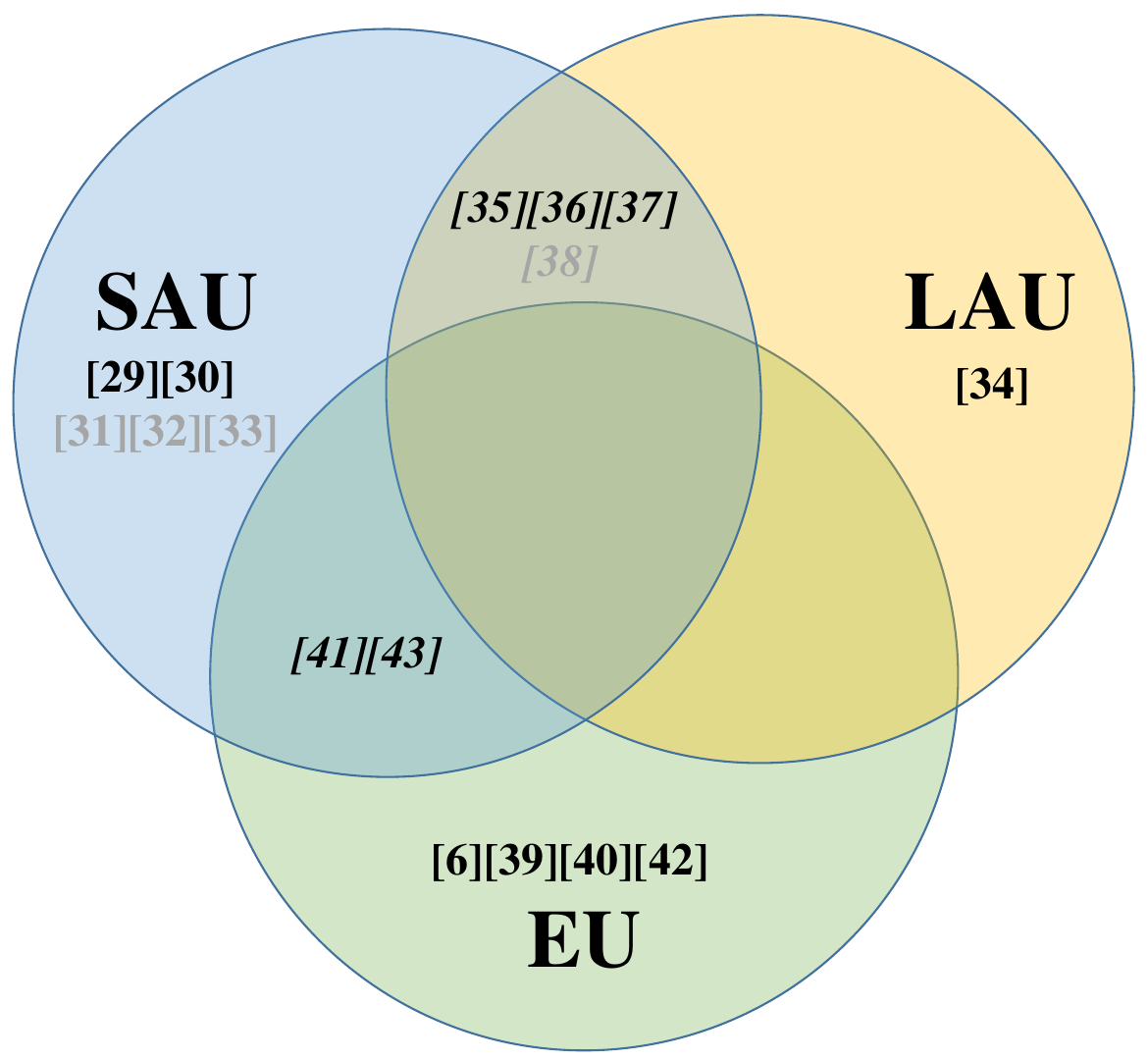}
    \caption{Overview of current UAP approaches: a classification based on the type of modeled uncertainty. \textcolor{gray}{[*]} indicates the absence of deep learning models, whereas \textit{[*]} denotes a lack of distinction among the various types of uncertainties modeled or their effects.}
    \label{fig:sup_01}
\end{figure}

Fig. \ref{fig:sup_01} encapsulates a summary of existing representative UAP methods in modeling and applying different uncertainties. There is a noticeable lack of simultaneous modeling and addressing of diverse uncertainties in the existing research. Moreover, there is an absence of systematic study and comparison within a unified UAP framework. This deficiency hinders the objective assessment of various uncertainties and risk-modeling approaches.
To address these gaps, this study aims to amalgamate, integrate, and compare these classes of uncertainty within a uniform framework. By thoroughly evaluating the impacts of SAU, LAU, EU, and their combinations on planning, this study  provides an in-depth understanding of their influence. This comprehensive approach aims to enhance the safety and dependability of autonomous systems and to foster the development of more robust and adaptable planning strategies. 

\section{Problem Formulation}

\begin{figure*}
    \centering
    \includegraphics[width=18cm]{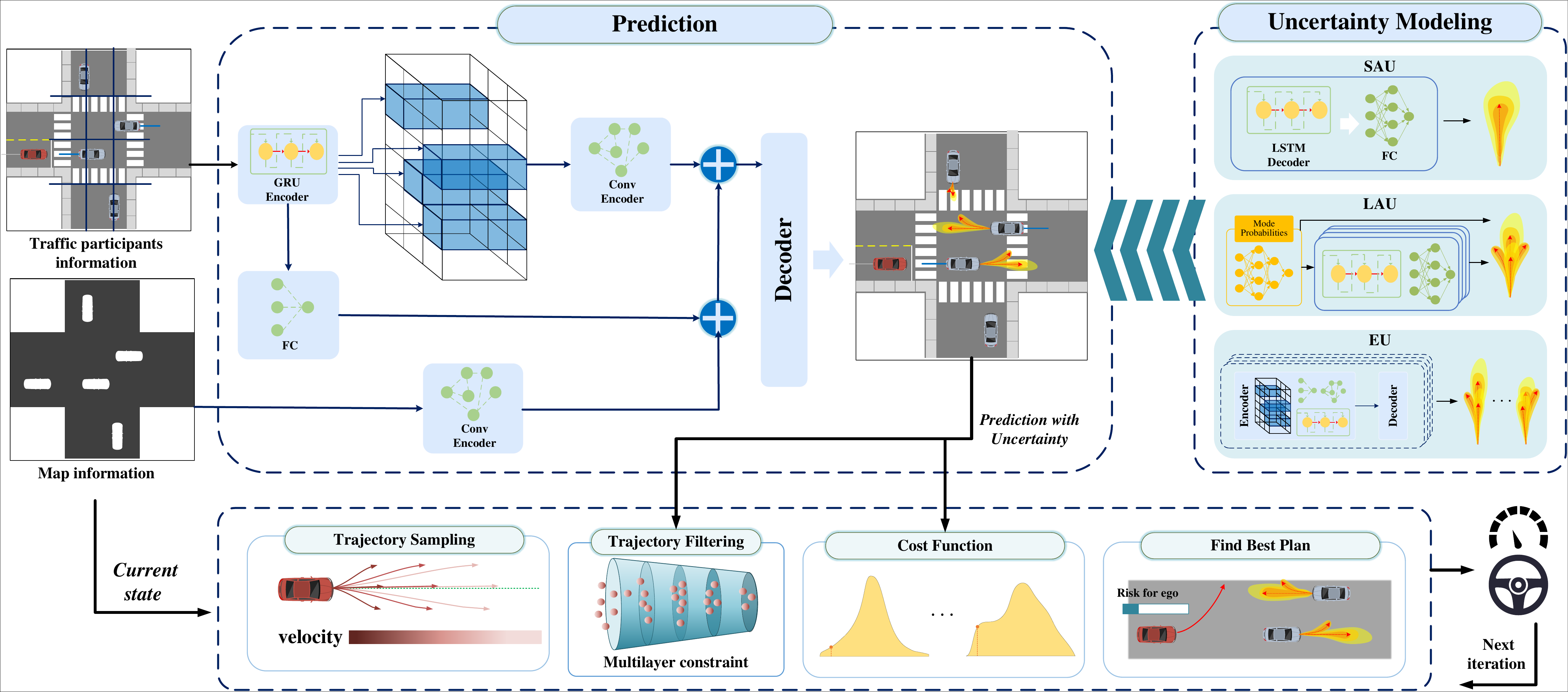}
    \caption{Proposed unified prediction and planning framework that considers different types of uncertainties.}
    \label{Fig2}
\end{figure*}

The ego vehicle \(A_{e}\) dynamically plans feasible trajectories by predicting the future states of other traffic participants \(A_{i}\), where \(i \in \{1, \ldots, N\}\). The ego vehicle's objective is to safely navigate to a target state by utilizing information about traffic participants and environmental factors, such as road configurations. This system is conceptualised as a continuous-space, discrete-time model, wherein each time step updates the states of traffic participants, encompassing their position, velocity, orientation, and shape.

The ego vehicle's navigation process begins by predicting the future motion $\mathbf{Y}=\{\mathbf{Y_{1:N}}\}$ of traffic participants $A_{1:N}$, based on spatiotemporal context \(\mathbf{I}_{pr}\). These predictions, modeled as probability distributions to encapsulate uncertainties, inform the planning phase, as follows:
\begin{equation}
    \mathbf{Y} \sim P(\mathbf{S},\mathbf{I}_{pr}),
\end{equation}
where \(\mathbf{S}=\{\mathbf{S_{1:N}}\}\) denotes the historical states of \(A_{i}\), including the position, velocity, and acceleration, alongside local map context. The goal is to estimate future trajectories \( \mathbf{Y_{i}}=\{\mathbf{y}_{i}^{1:t_{f}} \} \) for a time horizon \(t_f\), and the predicted trajectories are represented as \( \mathbf{\hat{Y}_{i}}=\{\mathbf{\hat{y}}_{i}^{1:t_{f}} \} \).

The planner, integrating prediction results and uncertainties, optimizes the ego vehicle's trajectory as follows:
\begin{equation}
\begin{aligned}
    \mathcal{T}_{\mathtt{opt}} &= \arg \min_{\mathcal{T}} C\left ( \mathcal{T}, P, \mathbf{I}_{pl} \right ), \\
    &\quad \mathcal{H}_1(\mathcal{T}, P, \mathbf{I}_{pl}) = 0, \\
    &\quad \mathcal{H}_2(\mathcal{T}, P, \mathbf{I}_{pl}) \leq 0,
\end{aligned}
\label{eq1}
\end{equation}

where \(\mathbf{I}_{pl}\) is the context information considered by the planner, such as the roads and global paths.
\(\mathcal{T}=\{\mathbf{s}_{e}^{1:t_{f}}\}\) denotes the planning trajectory, and the optimized trajectory is defined as \(\mathcal{T}_{\mathtt{opt}}\).
The main difficulty lies in formulating the cost function \(C\), particularly when calculating the risk cost \(C_{r}\), which is a key component of \(C\).
This study focuses on modeling a rational risk $Cr_{r}$ cost based on prediction with uncertainty $P$. 
Furthermore, the optimization problem is subjected to certain constraints, denoted as $\mathcal{H}_1$ and $\mathcal{H}_2$, encompassing fundamental dynamics and safety constraints. They help initially exclude trajectories that are clearly unsuitable.
Fig. \ref{Fig2} depicts the proposed prediction and planning frame work, integrating various uncertainties and risk models.

\section{Prediction with Multiple Uncertainty Estimations}
\subsection{Prediction Model Backbone}\label{sec:secIV.A}

Accurate prediction is critical for the safe execution of vehicle maneuvers and informed decision-making, which require advanced modeling techniques. The backbone of the prediction model in this study, finely adapted from a proven model \cite{geisslinger2021watch}, was tailored to capture complex dynamic interactions in traffic scenarios (Fig. \ref{Fig2}).
This model integrates various components for specific tasks. It employs GRU-based encoders for encoding historical state features of entities and combines these with convolutional social pooling layers to model interactions among different objects. The map features were processed using CNN layers and max-pooling. Thereafter, the encoded features are amalgamated, and a GRU-based decoder generates the final predictions. The network also incorporates activation functions, such as Leaky ReLU, to enhance information-processing efficiency.
Overall, this network represents a sophisticated model for trajectory prediction that can accommodate the complexities of real-world traffic. Its design, which features a combination of specialised layers and activation functions, provides a robust foundation for accurate prediction.
The deterministic prediction of future trajectories, without considering uncertainties, forms the baseline of this model. The following sections delve into the integration of various types of uncertainties into the prediction process.

\subsection{Aleatoric Uncertainty Estimation}
This section addresses the modeling of two types of AUs for trajectory prediction: SAU and LAU. Both were unifiedly modeled within the proposed prediction network.

\subsubsection{SAU}
It captures the immediate stochasticity in the motion of traffic participants, and noise originating from annotation or perception. The prediction with SAU is assumed to follow a unimodal Gaussian distribution. At each time, the position conforms to a bivariate Gaussian distribution $P_{SAU} \mathrel{\mathop:}= \mathcal{N}(\boldsymbol{\mu}^{t}, \boldsymbol{\Sigma}^{t})$. The expected position $\boldsymbol{\mu}^{t} = (\mu_x^{t}, \mu_y^{t})$ and covariance matrix $\boldsymbol{\Sigma}^{t}$ at each time are simultaneously modeled in the network output.

\subsubsection{LAU}
This arises from the multimodal motion of traffic participants. For example, a vehicle may accelerate, decelerate, turn left, or turn right. To enhance its applicability across diverse scenarios, this study refrains from explicitly modeling or constraining specific behaviour patterns. Instead, it adopts the form of a GMM to capture this multimodality, allowing it to adaptively learn optimal modal features and their distributions based on real-world data. At each time step:

\begin{equation}
    \mathbf{y} \sim P_{AU} \mathrel{\mathop:}= \sum_{k=1}^{K} w_k \cdot \mathcal{N}(\boldsymbol{\mu}_k, \boldsymbol{\Sigma}_k).
\label{eq2}
\end{equation}
The decoder in Section \ref{sec:secIV.A} first infers the weights of different modes. Subsequently, each mode is encoded as a one-hot vector, which is then combined with the encoded features to estimate the Gaussian distribution for each mode.

\subsubsection{Training process for AU}
A customised training regimen was designed to enable the model to accurately forecast trajectories, while accounting for uncertainties. It consists of two distinct phases. In the initial phase, the weighted mean squared error (wMSE) loss function is employed to guide the model in generating precise projected trajectories.

\begin{equation}
    L_{mse}=\sum_{k=1}^{K} w_k\|\mathbf{Y}-\mathbf{\hat{Y}_{k}}\|_{2}^{2}
\label{eq2.5}
\end{equation}

In the second phase, the model is guided using weighted negative log likelihood (wNLL) loss to ensure rational uncertainty estimates.

\begin{equation}
    L_{nll}= -\log (P(\mathbf{Y}|\mathbf{S}))=-\sum_{k=1}^{K} w_k \sum_{t=1}^{T} \log (\mathcal{N}_{k}^{t}(\mathbf{y}^{t}|\mathbf{S}))
\label{eq3}
\end{equation}

Notably, in certain existing multimodal trajectory prediction algorithms, a 'winner-takes-all (WTA)' training strategy is frequently adopted \cite{deo2022multimodal}. However, focusing solely on the best trajectory was considered inadequate. For instance, insufficient prior knowledge to guide the decision model in selecting the optimal predicted trajectory can lead to suboptimal outcomes. Conversely, neglecting other potentially less favourable predicted trajectories may significantly compromise planning performance. Therefore, this study employs a weighted approach to compute a comprehensive loss by placing a duphaseis on the predicted trajectories of the different modes. The learned weights also guides the planning process.

This approach addresses both SAU and LAU in trajectory prediction. By employing GMM, the model demonstrates versatility in capturing multimodal behaviour patterns. The two-stage training process ensures a balanced consideration of the accuracy and uncertainty estimation, ultimately enhancing the predictive capabilities of the model.

\subsection{Epistemic Uncertainty Estimation}
Well-designed models have the potential to exhibit a high EU when faced with unfamiliar scenarios. This attribute facilitates reliable planning by safeguarding against situations that deviate from the training data.
Deep ensemble has gained widespread recognition as an effective, flexible, and versatile approach for estimating EU. In this study, the design and application of deep ensemble within a prediction-planning pipeline were investigated.

The implementation of deep ensemble involves two crucial operations: bootstrapping and random initialisation of models. For a given training set \(D\), \(M\) sets of training subsets \(\{D_{1}, \ldots, D_{M}\}\) of equal size are drawn through a random sampling process with replacement. These subsets are then employed to train \(M\) distinct but structurally identical submodels with parameters $\theta_{1}, ..., \theta_{M}$. It is imperative to emphasize that each submodel undergoes an individualized random initialization preceding its training, whereas uniformity is maintained in subsequent stages.

During the inference phase, the input undergoes forward propagation through all submodels, yielding \(M\) predicted results \(P_{\theta_{1}}, ..., P_{\theta_{M}}\), where each result can be either \(P_{SAU}\) or \(P_{AU}\). The divergence observed in the predictions among all submodels serves as a measure of EU. While various metrics, such as variance or entropy, can be used to quantify this uncertainty in the prediction \cite{shao2023failure}, this study specifically explores its application in the planning process. Therefore, multiple sets of predictions are retained, ensuring that the decision module has access to a comprehensive view of potential trajectories, thereby enhancing the adaptability and safety of the model in real-world traffic scenarios.

In summary, different types of uncertainty in prediction can be modeled separately within a unified framework, as required. By aggregating all the extracted uncertainties, a comprehensive prediction \(P\) containing the uncertainty can be obtained.

\section{Uncertainty-Aware Planning}
\subsection{Cost Function}\label{sec:cost_fun}
Autonomous driving motion planning is a typical multi-objective optimization problem. Therefore, the cost function in Eq. (\ref{eq1}) generally includes multiple components:

\begin{equation}
    C = k_{r}C_{r} + k_{b}C_{b} + k_{t}C_t + k_{g}C_{g},
\label{eq4}
\end{equation}
where \(C_{r}\), \(C_{b}\), \(C_{t}\), and \(C_{g}\) represent the costs related to risk from other traffic participants, boundary violations, target deviations, and global path deviations, respectively. \(k_{r}\), \(k_{b}\), \(k_{t}\), and \(k_{g}\) denote the corresponding weightings, which were manually set. \(C_{r}\) primarily depends on the interaction between the planned trajectory of $A_{e}$ and the future motion of $A_{1:N}$. This will be elaborated upon in detail in the following section. \(C_{b}\) depends on whether the ego vehicle collides with the road boundaries. This cost was assessed based on the potential harm resulting from such collisions. \(C_{t}\) is influenced by the deviation between the current state and the planned target. Following \cite{geisslinger2023ethical}, this cost was estimated by comparing the expected with planned speeds. \(C_{g}\) is calculated based on the deviation between the planned trajectory and the global path. 
This multi-component cost provides a comprehensive framework for addressing various objectives in the motion planning of autonomous driving.

\subsection{Uncertainty-Aware Risk Model}
\begin{figure}
    \centering
    \includegraphics[width=9cm]{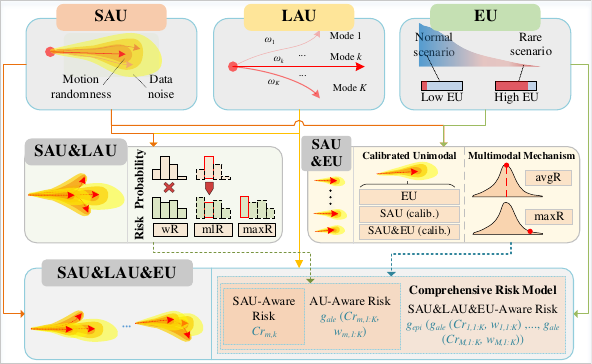}
    \caption{The modeled uncertainties and their combinations, as well as various uncertainty-aware risk models.}
    \label{fig3}
\end{figure}

The planned trajectories should avoid conflicts with the future motions of the surrounding traffic participants. This forms the basis for constructing the risk-cost term, \(C_{r}\). Assuming the future motion states of traffic participants \(A_{i}\) are known as \(\{\mathbf{s}_{i}^{1:t_{f}} \}\), and combining with the shape of \(A_{i}\), its occupied space over the next \(t_{f}\) time steps
can be computed. Correspondingly, based on the planned trajectory \(\mathcal{T}\) of ego vehicle \(A_{e}\) and its shape, its occupied space over the next \(t_{f}\) time steps
is calculated. The objective of introducing risk cost is to maintain a safe distance between the future occupied spaces of $A_{e}$ and $A_{1:N}$.

During planning, the predictor estimated the future motion of $A_{1:N}$. However, owing to the complexity of the environment, limited data, and model constraints, it is challenging to ensure the absolute accuracy of the predicted trajectories. By incorporating the uncertainty estimated in Section IV, the planner can adaptively adjust the risk model \(C_r = \phi(\mathcal{T}, P) \), thereby generating reasonable trajectories.
Fig. \ref{fig3} illustrates the uncertainties and their combinations, as well as the various risk models derived.

\subsubsection{SAU-Aware Risk Model}
Risk is typically defined as the product of collision probability and harm. When considering only SAU, the collision risk between the ego vehicle \(A_{e}\) and traffic participant \(A_{i}\) can be calculated as

\begin{equation}
    C_{r}=\sum_{i=0}^{N}\max_{1:t_f}(P_{c_{i}}^{t} \times H_{i}^{t}),
\label{eq5}
\end{equation}
where \(H_{i}^{t}\) represents the harm if a collision occurs between \(A_{e}\) and \(A_{i}\) at time \(t\). This is determined by parameters such as the mass of the objects involved, relative velocity, and collision angle. This study adopts the harm model from \cite{geisslinger2023ethical} by selecting the unimodal Gaussian distribution peak predicted at each time as the state of \(A_{i}\).
\(P_{c_{i}}^{t}\) represents the predicted collision probability between \(A_{e}\) and \(A_{i}\) at time \(t\). It is computed based on the predicted probability distribution of \(A_{i}\), planned trajectory of $A_{e}$, and their respective shapes.

SAU was modeled as a continuous probability distribution at each time. Therefore, the probability density function (PDF) \(f(x_{i}^{t}, y_{i}^{t} \mid \boldsymbol{\mu}_{i}^{t}, \boldsymbol{\Sigma}_{i}^{t})\) of the predicted trajectory at time \(t\) is used to calculate the collision probability \(P_{c_{i}}^{t}\). Assuming the distribution considering the shape of \(A_{i}\) is \(f_{s}(x_{i}^{t}, y_{i}^{t} \mid \boldsymbol{\mu}_{i}^{t}, \boldsymbol{\Sigma}_{i}^{t})\), $P_{c_{i}}^{t}$ is obtained by integrating \(f_{s}\) over the space relevant to the ego vehicle:

\begin{equation}
    P_{c_{i}}^{t}=\iint_{S_{e}^{t}} f_s(x_{i}^{t}, y_{i}^{t} \mid \boldsymbol{\mu}_{i}^{t}, \boldsymbol{\Sigma}_{i}^{t}) d {S_{e}^{t}},
\label{eq6}
\end{equation}
where \(S_{e}^{t}\) represents the space occupied by \(A_{e}\) at time \(t\), which is determined by the planned trajectory and the shape of $A_{e}$. For computational efficiency, an approximate model was employed to replace the above formula to obtain the probability distribution \(f_{s}\) considering the shape of \(A_{i}\). The midpoints of the front, middle, and rear ends were selected as the centres of the three bivariate Gaussian distributions. The estimated \(\boldsymbol{\Sigma}_{i}^{t}\) is simultaneously used as the covariance of the three distributions. The integrals of the three distributions over \(S_{e}^{t}\) are averaged to obtain the final collision probability. Additionally, for the ego vehicle \(A_{e}\), it is divided into three small rectangles along its long side. Thereafter, they are replaced by equally sized rectangles along the x-axis, thus approximating \(S_{e}^{t}\). The risk cost, which is sensitive to SAU, can be obtained based on the estimated results of collision probability and harm. Furthermore, the above risk model will serve as a basic element for further exploration of LAU and EU modeling.

\subsubsection{AU-Aware Risk Model}\label{sec:LAU_risk}

LAU is modeled as a multimodal probability distribution that incorporates weights. Compared to modeling only SAU, simultaneously modeling both types of AU takes the form of a multipeaked Gaussian distribution. A new modeling approach is required to account for the influence of LAU. The risk is formulated by
\begin{equation}
    C_r = f_r(\mathcal{T}, \sum_{k=1}^{K} w_k \cdot \mathcal{N}(\boldsymbol{\mu}_k, \boldsymbol{\Sigma}_k)) =g_{ale}(C_{r_{1}}...C_{r_{K}}, w_{1}...w_{K}),
\label{eq7}
\end{equation}
where \(C_{r_{i}}\) represents the risk generated by the distribution corresponding to the \(i\)-th mode, obtainable using Eq. (\ref{eq5}). By treating the multipeaked Gaussian distribution as multiple unimodal Gaussian distributions, it can be integrated with the aforementioned risk model. \(g_{ale}()\) represents the modeling method for the multimodal risk, which can be categorized into:

\begin{itemize}
    \item \textbf{Weighted Risk (wR)}: It models the risks generated by all modes under the predicted distributions, and the final output is a composite risk value.
    \begin{equation}
        C_r = \sum_{k=1}^{K} w_k C_{r_{k}}
    \label{eq8}
    \end{equation}

    \item \textbf{Most Likely Mode Risk (mlR)}: It only considers the risk generated by the most likely mode.
    
    \begin{equation}
        C_r = C_{r_{k'}},  k'=\arg\max_{k\in\{1,...,K\}} w_{k}
    \label{eq9}
    \end{equation}
    
    \item \textbf{Maximum Risk (maxR)}: It only considers the maximum risk among all modes, representing the worst-case situation. This is a conservative modeling approach.
    
    \begin{equation}
        C_r = \max_{k\in\{1,...,K\}} C_{r_{k}}
    \label{eq10}
    \end{equation}
\end{itemize}

The three modeling approaches delineate distinct cognitive inclinations toward risks induced by LAU. The subsequent sections will delve into the application of these models during planning, followed by a comprehensive comparative and analytical discussion within the experimental section.

\subsubsection{EU-Aware Risk Model}\label{sec:EU_risk}

The acquisition of EU arises from the integration of the forward-propagated outcomes of \(M\) submodels. The corresponding risk model is elucidated as

\begin{equation}
    C_r = f_{r}(\mathcal{T}, \{P_{\theta_{1}}, ..., P_{\theta_{M}}\}),
\label{eq11}
\end{equation}
where \(P_{\theta_{m}}\) represents the prediction result of the \(m\)-th submodel. To simplify representation, assume \(P_{\theta_{m}}\) represents a Gaussian distribution of an object predicted at a specific moment. This section primarily focuses on the case when \(P_{\theta_{m}}\) only models SAU. To model the influence of EU, two distinct modeling approaches are designed: risk modeling based on the integration of $M$ unimodal distributions, and risk modeling based on a multimodal mechanism.

\textbf{Risk Modeling Based on the Calibrated Unimodal Distribution}: A calibrated unimodal Gaussian distribution \(P(x,y|\boldsymbol{\mu}, \boldsymbol{\Sigma})\) can be constructed based on both SAU and EU. It can then be further used to compute the calibrated risk using Eq. (\ref{eq5}). Assuming that the parameters of \(P_{\theta_{m}}\) include the mean \(\boldsymbol{\mu_{m}}\) and covariance \(\boldsymbol{\Sigma_{m}}\), the total uncertainty, considering both EU and SAU, is obtained by integrating the distribution outputs of all the submodels. The calibrated mean is expressed as follows:

\begin{equation}
    \boldsymbol{\mu} = \frac{1}{M}\sum_{m=1}^{M}{\boldsymbol{\mu_{m}}}.
\label{eq12}
\end{equation}

Additionally, the covariance matrix corresponding to the total uncertainty is

\begin{equation}
    \boldsymbol{\Sigma} = \boldsymbol{E}_{\boldsymbol{\mu}\boldsymbol{\mu}^{T}} - \boldsymbol{\mu}\boldsymbol{\mu}^{T},
\label{eq13}
\end{equation}
where \(\boldsymbol{E}_{\boldsymbol{\mu}\boldsymbol{\mu}^{T}}\) represents the second moment, calculated as

\begin{equation}
    \boldsymbol{E}_{\boldsymbol{\mu}\boldsymbol{\mu}^{T}} = \frac{1}{M} \sum_{m=1}^{M} {\boldsymbol{\Sigma}_{m} + \boldsymbol{\mu}_{m}\boldsymbol{\mu}_{m}^{T}}.
\label{eq14}
\end{equation}

For comparison, two covariance matrix estimation methods were designed to consider only the calibrated AU or EU. The calibrated SAU is calculated as

\begin{equation}
    \bar{\boldsymbol{\Sigma}} = \frac{1}{M} \sum_{m=1}^{M}\boldsymbol{\Sigma_{m}}.
\label{eq15}
\end{equation}

To obtain the distribution considering only EU, the means \(\mu_{1:M}\) of \(M\) distributions are extracted to form a set of points to fit a new Gaussian distribution. The correlation coefficient of this distribution is assumed to be 0, indicating the independence between the \(x\) and \(y\) directions. This facilitates the computation of the means and variances for the \(M\) points, thus enabling estimation of distributions solely arising from EU.

\textbf{Risk Modeling Based on Multimodal Mechanism}: Under this mechanism, the outputs of multiple submodels are integrated into a GMM, where each mode's weight is set to \(1/M\). Similar to the treatment of LAU, the composite risk \(C_r\) is calculated based on the risks corresponding to each mode (i.e., submodel), which includes two types of risk preferences:

\begin{itemize}
    \item \textbf{Average Risk (avgR)}: It averages the risks from all modes, which is beneficial in mitigating the problem of inaccurate prediction due to the failure of a single model. Such a failure may occur when the model is insufficiently trained or when it encounters an unfamiliar scenario.
    \item \textbf{Maximum Risk (maxR)}: It emphasizes the worst-case scenario, effectively handling situations where one or more models might be excessively optimistic in risk assessment. Considering this worst-case risk, the planner is prompted to formulate the trajectory for the ego vehicle in a conservative manner.
\end{itemize}

\subsubsection{Comprehensive Risk Model}\label{sec:comprehinsive risk}

While the preceding risk models effectively quantify up to two types of uncertainties, there remains a need to address all three concurrently. In this section, we propose a comprehensive hierarchical risk model to rectify this limitation, as follows:

\begin{equation}
    C_{r} = g_{epi}(g_{ale}(C_{r_{1,1:K}}, w_{1,1:K}),...,g_{ale}(C_{r_{M,1:K}, w_{M,1:K}})).
\label{eq16}
\end{equation}

This model comprises three tiers: \(C_{r_{m,k}}\) represents the first level, tasked with quantifying the risk under SAU associated with the \(k\)-th mode of the \(m\)-th submodel. The \(g_{ale}\) operation forms the second-level operation aimed at integrating the risks of all modes from a single submodel, thus accounting for LAU, as detailed in Section \ref{sec:LAU_risk}. The \(g_{epi}\) operation serves as the third-level operation, facilitating the amalgamation of results from various submodels to account for EU, as detailed in the design of risk modeling based on the multimodal mechanism in Section \ref{sec:EU_risk}. Algorithm \ref{Algorithm1} outlines the construction process of the comprehensive risk model, with the resulting comprehensive risk serving as a crucial input to the planner.

\begin{algorithm}
\caption{Calculation of Comprehensive Risk}
\label{Algorithm1}
\KwIn{Training dataset $\mathcal{D}$, candidate planning trajectories \(\mathcal{T}\), historical states $\mathbf{S}$ of traffic participants, and context.}
\KwOut{Estimated comprehensive risk $C_{r}$, considering SAU, LAU, and EU.}

\textbf{Training Phase:} \\
Sample $M$ training subsets from $\mathcal{D}$ using bootstrapping\;
Randomly initialize $M$ prediction networks (modeling GMM for both SAU and LAU)\;
Train $M$ models with these subsets and networks\;

\textbf{Inference Phase:} \\
Given states of traffic participants and context, predict $M$ sets of distributions $P_{\theta_{1}},\ldots,P_{\theta_{M}}$ in parallel\;
\For{$P_{\theta_{m}}$ in the predicted distributions}{
    \For{$P_{\theta_{m,k}}$ in all modes of $P_{\theta_{m}}$}{
        Calculate risk for $P_{\theta_{m,k}}$ using Eq. (\ref{eq6}) and (\ref{eq7})\;
    }
    Calculate risk $g_{ale}(C_{r_{m,1:K}}, w_{m,1:K})$ for $P_{\theta_{m}}$ using Eq. (\ref{eq8}) to (\ref{eq10})\;
}
Calculate comprehensive risk $C_{r}$ using Eq. (\ref{eq16})\;
\end{algorithm}


\subsection{Uncertainty-Aware Planning}
UAP is an integral aspect of the proposed methodology and seamlessly complements the preceding uncertainty-based risk modeling. They play a pivotal role in enabling vehicles to make informed decisions in intricate and dynamic environments. This process encompasses two steps: candidate trajectory generation, and optimal trajectory selection.

\subsubsection{Candidate Trajectory Generation}

\begin{figure}
    \centering
    \includegraphics[width=9cm]{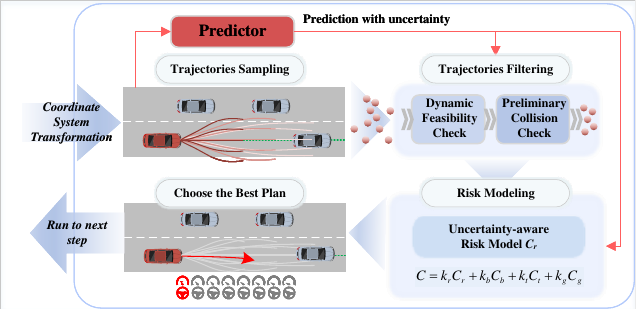}
    \caption{The process of uncertainty-aware planning.}
    \label{fig5}
\end{figure}

At each time step, the planner generates a candidate trajectory set \(\mathcal{\mathbf{T}} = \{\mathcal{T}_{1},..., \mathcal{T}_{P}\}\), where each candidate trajectory \(\mathcal{T}_{i}\) comprises the future planned states of the ego vehicle, which include the vehicle’s position, velocity, acceleration, and orientation.

This study adopted the Frenet coordinate system, which is a specialised local framework designed for autonomous driving scenarios. This system characterises a vehicle's position and orientation relative to a reference path. It comprises two essential components: the longitudinal coordinate \(s\), representing distance along the reference path, and the lateral coordinate \(d\), signifying lateral deviation from the reference path. The \(s\)-axis is aligned with the direction of the path, whereas the \(d\)-axis is perpendicular.

Following the acquisition of planning objectives, an initial global path was constructed to serve as the reference path. At each time step, a new set of candidate trajectories is generated. Candidate trajectory generation based on the Frenet coordinate system aims to find feasible trajectories adhering to specific constraints and optimization criteria by exploring various combinations of longitudinal and lateral motions. It employs a polynomial curve for candidate trajectory construction: fourth- and fifth-order polynomials for longitudinal and lateral motions, respectively. The former enables smooth and continuous changes in the longitudinal position, velocity, and acceleration, capturing the complex dynamics of the vehicle's motion and enabling precise navigation through the environment. The latter provides the flexibility required to model the intricate lateral movements of the vehicle, allowing for smooth transitions between different lateral positions to ensure the feasibility of the generated trajectories.

Furthermore, for scenarios requiring emergency avoidance maneuvers, trajectories for emergency braking were generated, enabling evasion of obstacles located in front of the vehicle, as well as trajectories for emergency acceleration, facilitating the avoidance of obstacles situated behind the vehicle. These trajectories were designed to utilize the maximum deceleration and acceleration values within allowable ranges. All the aforementioned candidate trajectories constitute a comprehensive candidate trajectory set.

\subsubsection{Optimal Trajectory Selection}
The optimal trajectory \(\mathcal{T}_{\mathtt{opt}}\) was chosen from the pool of candidate trajectories. This is formulated as a constrained optimization problem, where the initial set of constraints pertains to dynamic limitations on velocity \(v\), acceleration \(a\), and curvature \(\kappa\). These constraints serve to sift through the candidate trajectory set, producing a fresh set of trajectories.

Following this, an initial collision check was employed to assess whether the vehicle might collide with either the road boundaries or other traffic participants. Collisions with road boundaries are determined by conflicts between candidate trajectories and fixed road boundaries.

The preliminary collision checks with other traffic participants involved extracting representative trajectories from the predicted probability distribution. In cases where only SAU is considered, the peak of the predicted distribution at each time step is extracted to form the checked trajectory, which is commonly referred to as the mean trajectory. When both LAU and SAU are considered, the mean trajectories from \(K\) modes are extracted following the same logic. The weighted collision rates are then computed and compared against a predefined threshold $CC_{th_{1}}$ to ascertain the check result.

For distributions considering both EU and SAU, mean trajectories from the output distributions of $M$ submodels are extracted. Two approaches are proposed: a collision check based on the integrated trajectory (\textit{ic.}) and check based on multimodal trajectories (\textit{mc.}). The former aggregates all mean trajectories by averaging to obtain an integrated trajectory, which is used for a collision check. The latter first obtains check results from all mean trajectories before calculating the average collision rate. This is compared with the threshold $CC_{th_{2}}$ to obtain the final check result.

A layered collision check approach was designed for distributions that simultaneously considered all three uncertainties. 
This entails computing collision check outcomes for the multimodal distributions generated by each submodel, and aggregating their outcomes. Following dynamic assessments, preliminary road boundary collision checks, and collision checks, the candidate trajectories were classified into distinct tiers. Thereafter, they were integrated with a cost evaluation to derive the optimal trajectory. Algorithm \ref{Algorithm2} delineates the comprehensive UAP framework where all three uncertainties are considered. Here, $CC_{i,m,k}$ denotes the collision check result based on the predicted mean trajectory from the $m$-th submodel for the $k$-th mode of $A_{i}$.

For each candidate trajectory, the cost function \(C\) constructed in Section \ref{sec:cost_fun} was used. The lowest-cost trajectory among those passing all checks was prioritized.
An autonomous vehicle operates continuously within a scene using a rolling optimization strategy. At each time step, the vehicle executes motion according to the resolved optimal trajectory, and subsequently re-plans based on the context information and prediction results for the next time step. This iterative process allows the vehicle to dynamically adapt to evolving traffic conditions and uncertainties, thereby ensuring that it navigates through complex scenarios with agility and precision. Additionally, the rolling optimization approach enables an autonomous vehicle to respond promptly to sudden changes in the environment, enhancing its overall safety and efficiency in real-world driving scenarios.

\begin{algorithm}
\caption{Uncertainty-Aware Planning}
\label{Algorithm2}
\SetAlgoNlRelativeSize{0}

\While{not meeting termination condition}{
    Generate candidate trajectories based on updated scene\;
    \ForEach{candidate trajectory $\mathcal{T}$}{
        Perform dynamic feasibility check\;
        \ForEach{traffic participant $A_{i}$}{
            Retrieve prediction outcomes for $A_{i}$\;
            \ForEach{prediction of each submodel}{
                \ForEach{mean trajectory of each mode}{
                    $CC_{i,m,k} \leftarrow \text{Collision check}$\;
                }
                $CC_{i,m} \leftarrow \sum_{k=1}^{K} w_{k} C_{i,m,k} > CC_{th_{1}}$\;
            }
            $CC_{i} \leftarrow \sum_{m=1}^{M} C_{i,m} > CC_{th_{2}}$\;
        }
        \If{there exists $C_{i} > 0$ for $i \in \{1, \ldots, N\}$}{
            Possible collision detected, check failed\;
        }
        Perform road collision check\;
    }
    Evaluate the cost by Algorithm \ref{Algorithm1}\;
    Prioritize selection of the lowest-cost trajectory among those that pass the checks\;
    Execute the first step of the optimal trajectory $\mathcal{T}_{\text{opt}}$\;
}
\end{algorithm}

\section{Experimental Setup}
\subsection{Experimental Platform and Data}

A comprehensive experimental platform (CommonRoad) and diverse traffic scenarios \cite{althoff2017commonroad} was employed to evaluate the proposed UAP methodologies under various uncertainties. The scenarios encompassed a blend of both real-world and simulated environments, including highways, intersections, and T-junctions, for a comprehensive and extensive evaluation of our methodology.

The dataset comprised 2000 scenarios partitioned into training, validation, and test sets in a 3:1:1 ratio. This partitioning strategy prevents model overfitting and allows an unbiased assessment of the real-world applicability of the model.

The evaluation utilized the CommonRoad benchmark, which enhances scenarios with planning challenges, vehicle dynamics, and parameters. Planning problems are defined by parameters, such as the target position, velocity, and time of arrival, offering a rigorous test of the proposed methods.
To achieve a comprehensive assessment of different uncertainties and UAP, the experiments were divided into two categories:

1. \textbf{Normal Testing:} Focusing on the common test scenarios, this evaluates the core performance of the methods, offering a direct and representative assessment.

2. \textbf{Testing under Limited Perception:} Acknowledging real-world perception limitations, occlusions or noise are introduced to mimic the effects caused by information gaps and distributional shifts. Thereafter, the model's performance in ensuring system robustness is evaluated.

\subsection{Evaluation Metrics}
The experiments involved assessing both the prediction module and planning system.
For the former, fundamental metrics including average displacement error (ADE), final displacement error (FDE), and negative log likelihood (NLL) were employed to evaluate unimodal predictions.
For the LAU-based multimodal prediction, weighted error metrics (including wADE, wFDE, and wNLL) were employed. Additionally, the Best-of-N (BoN) metric (including minADE, minFDE, and minNLL) was adopted to assess the multimodal prediction, thereby providing a more comprehensive assessment of the model's accuracy and reliability in capturing diverse and complex scenarios.

The integrated prediction-planning system is evaluated as

\begin{itemize}
    \item \textbf{SR (Success Rate):} Proportion of scenarios where the ego vehicle reaches the target state successfully.
    
    \item \textbf{CR (Collision Rate):} Proportion of scenarios with collisions events.
    
    \item \textbf{AS (Average Speed):} Average speed of the ego vehicle.
\end{itemize}

\subsection{Compared Methods}

To evaluate the prediction module, its performance under different uncertainty modeling approaches was compared.

For the evaluation of the prediction-planning system, the following methods were considered:

\begin{itemize}
    \item \textbf{Constant Velocity (CV)-based Prediction-Planning:} Using predicted trajectories from a CV model.
    
    \item \textbf{non-UAP:} Planning is based solely on the predicted deterministic trajectory, without considering any uncertainties in the prediction.
    \item \textbf{Trajectron++-based planning:} The representative Trajectron++ model \cite{salzmann2020trajectron++} was used for comparison. It predicts trajectories and models AU through a Conditional Variational Autoencoder (CVAE). Furthermore, we incorporate the proposed AU-aware risk models for planning.
    \item \textbf{Different varieties of UAP}.
    Within the proposed unified framework, a comprehensive comparison of various uncertainty-aware risk models, as depicted in Fig. \ref{fig3}, is conducted. This also includes implementations and comparisons of risk-modeling methods from existing representative works:
    \begin{itemize}
        \item \textbf{Modeling SAU only.}  The risk model from existing research\cite{geisslinger2023ethical} was implemented for comparison.
        \item \textbf{Modeling LAU only}. For comparison, the CVaR-based risk model from MARC\cite{li2023marc} was constructed as LAU (maxR).
        \item \textbf{Modeling EU only.} Based on representative studies, models based on EU (calib.)\cite{10184920} and EU (maxR)\cite{zhou2022long} are constructed for comparison.
        \item \textbf{Modeling SAU\&LAU simultaneously.} Various risk models proposed in Sect. \ref{sec:LAU_risk} are implemented.
        \item \textbf{Modeling SAU\&EU simultaneously.} SAU\&EU (calib.) is implemented as an existing method\cite{tang2022prediction} for comparison. Additionally, different risk models in Sec. \ref{sec:EU_risk} are implemented and discussed.
        \item \textbf{Modeling SAU\&LAU\&EU.} The comprehensive risk model in Sec. \ref{sec:comprehinsive risk} is implemented.
    \end{itemize}    
\end{itemize}

The aforementioned varieties represent uncertainties explicitly modeled in predictions. It is imperative to note that the selective modeling of certain uncertainties does not preclude the existence of other types. For instance, when modeling SAU only, the estimated uncertainty may also be influenced by LAU and EU owing to inherent modeling limitations. 
Furthermore, it is pertinent to clarify that our comparative analysis is confined to the risk modeling principles outlined in the various studies, intended for evaluation within a unified decision-making framework, rather than encompassing a comprehensive replication of the decision models themselves.

Additionally, this study compares different implementation methods for estimating each type of uncertainty:
\begin{itemize}
    \item For SAU estimation, a comparison is made with a single-phase training (SPT) based on $L_{nll}$\cite{zheng2021unlimited}.
    \item For LAU estimation, WTA strategy\cite{deo2022multimodal} is evaluated.
    \item For EU estimation, the bootstrap method is compared with existing data shuffling techniques\cite{lakshminarayanan2017simple,shao2023failure}.
\end{itemize}

\subsection{Implementation Details}
The prediction algorithm underwent 30 training epochs, comprising 20 and 10 epochs in the initial and subsequent stages, respectively.
Regarding the handling of training data, missing values are imputed in cases where the historical trajectory length is insufficient. For real future trajectories with inadequate lengths, the masking technique is applied to consider losses solely from valid trajectories.
In the GMM modeling process, \(K\) is set to four. In the implementation of the deep ensemble, \(M\) is set to five.
Candidate trajectory generation involved the creation of 225 primary and 16 emergency obstacle avoidance trajectories.
For deep ensemble, a comparative analysis was conducted to evaluate the impact of utilizing bootstrapping and varying \(M\) on the performance. This study substantiates the efficacy of the selected approach.

\section{Experimental Results and Discussion}

\subsection{Prediction Evaluation}
This section evaluates the prediction module performance, as listed in Table 1, where "SAU" uses unimodal distribution and "SAU\&LAU" applies GMM for multimodal distributions.    For "SAU\&EU," which combines unimodal distribution with deep ensemble, two evaluation approaches are used: 
one involves averaging trajectories based on multiple outputs to assess the ADE and FDE;  the other treats it as a multimodal distribution to evaluate the weighted error and the BoN metrics.
The prediction model employed here demonstrates superior performance over comparative models in terms of ADE and FDE metrics, with its advantages becoming even more pronounced in subsequent planning outcomes.

Incorporating LAU significantly increases the diversity and alignment of predicted trajectories with ground truth.    The favourable BoN performance show its advantages.
By incorporating an EU modeling based on deep ensemble techniques, we can directly improve predictions by obtaining more accurate average trajectories.  Moreover, modeling it as a multimodal distribution also enhances the diversity of predictions.
In contrast to randomly shuffling the training data, bootstrapping enhances the diversity between different submodels, resulting in improved BoN performance.


\begin{table*}[]
\caption{Evaluation of different prediction methods. \textit{w/o boot.} indicates that data is not bootstrapped but rather randomly shuffled.}
\centering
\begin{tabular}{m{3.0cm}<{\centering} | m{1.2cm}<{\centering}  m{1.2cm}<{\centering}  m{1.2cm}<{\centering} | m{1.2cm}<{\centering}  m{1.2cm}<{\centering}  m{1.2cm}<{\centering} | m{1.2cm}<{\centering}  m{1.2cm}<{\centering}  m{1.2cm}<{\centering}}
\toprule
                  & \textbf{ADE}   & \textbf{FDE}   & \textbf{NLL}    & \textbf{wADE}  & \textbf{wFDE}  & \textbf{wNLL}  & \textbf{minADE} & \textbf{minFDE} & \textbf{minNLL} \\
\midrule
\textbf{CV}                & 1.638 & 3.962 & -      & -     & -     & -     & -      & -      & -      \\
\textbf{Trajectron++}           & -     & -     & -      & 1.303 & 3.624 & -1.091 & 0.689  & 1.847  & -3.847  \\
\midrule

\textbf{SAU}               & 0.805 & 2.129 & -0.621  & -     & -     & -     & -      & -      & -      \\
\textbf{SAU\&LAU}           & -     & -     & -      & 0.834 & 2.146 & -0.364 & 0.745  & 1.715  & -0.439  \\
\textbf{SAU\&EU (\textit{w/o boot.})} & 0.705 & 1.944 & - & 0.806     & 2.154     & -0.611     & 0.469  & 1.168  & -1.374 \\
\textbf{SAU\&EU}            & 0.665 & 1.805  & -  & 0.789  & 2.070     & -0.044     & 0.431  & 0.984  & -1.594 \\ 
\bottomrule
\end{tabular}
\label{Tab1}
\end{table*}

\subsection{Comparison of Different Methods}
This section provides a comparison of different uncertainty modeling approaches and UAP. Methods that modeled SAU, LAU, EU individually and combined were evaluated. The goal is to understand how different modeling strategies impact the overall performance of prediction and planning systems.

\subsubsection{Planning under Aleatoric Uncertainty}

Table \ref{Tab2} shows that incorporating AU estimation in prediction and planning notably enhances decision-making performance, by comparing rows 2–3 and 14–16. 
For example, compared with non-UAP, modeling only SAU or LAU results in a 5.3\% SR increase. 

Furthermore, jointly modeling SAU and LAU leads to greater improvements than individual modeling. This proves SAU\&LAU depicts a rational diversity output with a more rational characterisation of driving risks.
Contrastingly, maxR, as a conservative risk principle for worst-case situations, exhibit the best performance. In comparison to modeling SAU alone, it not only captures distributions closer to the ground truth, but also captures the mode more conducive to planning reasonable trajectories (3.3\% improvement in SR). This efficiently reduces collisions while operating effectively.

Additionally, compared to Trajectron++-based planning (refer to rows 4–6), the multimodal prediction network designed in this study resulted in improved planning performance (refer to rows 14–16), highlighting the advantages of the proposed prediction-planning method.

The comparison of rows 4–5 shows that the two-stage training strategy significantly outperformed SPT. This validates prioritizing the distribution mean optimization being performed prior to that of the entire distribution. Rows 6–8 show that the losses based on wMSE and wNLL achieves a more rational estimation of LAU compared with the WTA strategy, particularly in terms of enhancing plannning performance.

\begin{table}[]
\caption{Comparation of Different Planning Methods.}
\centering
\begin{tabular}{m{3.2cm}<{\centering}  m{1.2cm}<{\centering}  m{1.2cm}<{\centering}  m{1.2cm}<{\centering}}
\toprule
 & \textbf{SR} & \textbf{CR} & \textbf{AS} \\
\midrule
\textbf{CV}             & 0.732   & 0.260     & 5.498              \\
\textbf{non-UAP}       & 0.786   & 0.206     & 5.554              \\
\midrule
\textbf{Trajectron++ (wR)}            & 0.740   & 0.253     & 5.556              \\
\textbf{Trajectron++ (mlR)}            & 0.734   & 0.258     & 5.697              \\
\textbf{Trajectron++ (maxR)}            & 0.797   & 0.195     & 5.600              \\

\midrule
\textbf{SAU (SPT)}            & 0.760   & 0.232     & 5.464              \\
\textbf{SAU\&LAU (SPT)(maxR)}  & 0.560   & 0.435     & 5.257              \\
\midrule
\textbf{SAU\&LAU (WTA)(wR)}  & 0.841   & 0.148     & 5.688              \\
\textbf{SAU\&LAU (WTA)(mlR)}  & 0.812   & 0.177     & 5.604              \\
\textbf{SAU\&LAU (WTA)(maxR)}  & 0.841   & 0.148     & 5.695              \\
\midrule
\textbf{SAU}            & 0.839   & 0.154     & 5.656              \\
\textbf{LAU (maxR)}            & 0.839   & 0.154     & 5.700              \\
\midrule
\textbf{SAU\&LAU (wR)}   & 0.852   & 0.141     & 5.674              \\
\textbf{SAU\&LAU (mlR)}  & 0.849   & 0.143     & 5.674              \\
\textbf{SAU\&LAU (maxR)} & \textbf{0.867}   & \textbf{0.125}     & \textbf{5.801}               \\ 
\bottomrule
\end{tabular}
\label{Tab2}
\end{table}

\subsubsection{Planning with Consideration of Epistemic Uncertainty}
In this section, the focus is on planning considerations for EU. Given its potential to reflect underlying model limitations or imperfect knowledge, this section explores the practical performance gains when applying it to the prediction-planning system. The performances of the different EU-aware methods are listed in Table \ref{Tab3}. SAU (calib.) represents the AU calibrated through deep ensemble, and SAU\&EU (calib.) denotes the total uncertainty calibrated.

Solely modeling EU while ignoring SAU proves insufficient, as shown when comparing rows 2–4. This highlights the importance of a foundational SAU estimation in harnessing the benefits of EU.
Additionally, as an uncertainty calibration method, the deep ensemble yielded performance gains. This is shown in rows 4–5, further affirming the advantages of modeling EU in planning.
However, from rows 6–7, constructing the risk model by treating the results modeled based on SAU\&EU as a multimodal distribution leads to superior outcomes. This is particularly prominent when multimodal trajectories are employed in preliminary collision checks (\textit{mc.}). The subsequent text defaults on the adoption of (\textit{mc.}) approach.

Risk modeling based on maxR shows excellent performance, with a 5.5\% improvement in SR compared to modeling only SAU. This aligns with the earlier conclusion regarding modeling SAU\&LAU, emphasizing that planning to consider worst-case situations can enhance driving safety while maintaining efficiency.

\begin{table}[]
\caption{Evaluation of Different UAP Strategies Considering EU.}
\centering
\begin{tabular}{m{3.0cm}<{\centering}  m{1.2cm}<{\centering}  m{1.2cm}<{\centering}  m{1.2cm}<{\centering}}
\toprule
 & \textbf{SR} & \textbf{CR} & \textbf{AS} \\
\midrule
\textbf{SAU}                  & 0.839   & 0.154     & 5.656              \\
\textbf{EU (calib.)}                   & 0.823   & 0.169     & 5.620              \\
\textbf{EU (maxR)}                   & 0.839   & 0.154     & 5.645              \\
\midrule
\textbf{SAU (calib.)}            & 0.849   & 0.143     & 5.654              \\
\textbf{SAU\&EU (calib.)} & 0.846   & 0.146     & 5.652              \\
\midrule
\textbf{SAU\&EU (avgR)(\textit{ic.})}  & 0.859   & 0.133     & 5.709              \\
\textbf{SAU\&EU (maxR)(\textit{ic.})}   & 0.854   & 0.138     & 5.738              \\
\textbf{SAU\&EU (avgR)(\textit{mc.})}  & 0.875   & 0.117     & 5.724              \\
\textbf{SAU\&EU (maxR)(\textit{mc.})}   & \textbf{0.885}   & \textbf{0.107}     & \textbf{5.754}                  \\ 
\bottomrule
\end{tabular}
\label{Tab3}
\end{table}

Comparing Table \ref{Tab3} and \ref{Tab4} illustrates that bootstrapping leads to a reduction in the richness of data acquired by individual submodels during the training phase; therefore, it is not the optimal approach for training a single model. However, this differentiated treatment of the training set results in stronger diversity among the different submodels. This inclination towards optimizing for more dispersed local optima enables the integrated model to exhibit a robust adaptability to unfamiliar scenarios, ultimately leading to an improvement in UAP performance.

Parallel computations and model pruning are effective strategies for mitigating the increase in computational cost from multiple inferences. However, EU should be estimated with as few submodels as possible. Table \ref{Tab5} provides a comparison of UAP under different \(M\). It is evident that constructing only two submodels can achieve results comparable to modeling SAU alone. The setting \(M=5\) emerges as a well-balanced and optimal choice.

\begin{table}[]
\caption{Results when Modeling the EU with Random Shuffle of Data.}
\centering
\begin{tabular}{m{3.6cm}<{\centering}  m{1cm}<{\centering}  m{1cm}<{\centering}  m{1cm}<{\centering}}
\toprule
 & \textbf{SR} & \textbf{CR} & \textbf{AS} \\
\midrule
\textbf{SAU\&EU (calib.)(\textit{w/o boot.})} & 0.841   & 0.151     & 5.641              \\
\textbf{SAU\&EU (avgR)(\textit{w/o boot.})}  & 0.865   & 0.128     & 5.743              \\
\textbf{SAU\&EU (maxR)(\textit{w/o boot.})}   & 0.870   & 0.122     & 5.737              \\
\bottomrule
\end{tabular}
\label{Tab4}
\end{table}

\begin{table}[]
\caption{Evaluation of SAU\&EU (maxR) based UAP with various $M$.}
\centering
\begin{tabular}{m{3cm}<{\centering}  m{1.2cm}<{\centering}  m{1.2cm}<{\centering}  m{1.2cm}<{\centering}}
\toprule
 & \textbf{SR} & \textbf{CR} & \textbf{AS} \\
\midrule
\textbf{M=2}    & 0.852   & 0.141     & 5.688              \\
\textbf{M=3}    & 0.859   & 0.133     & 5.759              \\
\textbf{M=4}    & 0.872   & 0.120     & 5.732              \\
\textbf{M=5}    & \textbf{0.885}   & \textbf{0.107}     & 5.754              \\
\textbf{M=6}    & 0.867   & 0.125     & 5.731              \\
\textbf{M=7}    & 0.875   & 0.117     & 5.756              \\
\textbf{M=8}    & 0.872   & 0.120     & 5.770              \\
\textbf{M=9}    & 0.883   & 0.109     & 5.778              \\
\textbf{M=10}   & 0.875   & 0.115     & \textbf{5.782}               \\ 
\bottomrule
\end{tabular}
\label{Tab5}
\end{table}

\begin{table}[]
\caption{Comprehensive Comparison of UAP Modeling Based on Different Types of Uncertainties.}
\centering
\begin{tabular}{m{3.2cm}<{\centering}  m{1.2cm}<{\centering}  m{1.2cm}<{\centering}  m{1.2cm}<{\centering}}
\toprule
 & \textbf{SR} & \textbf{CR} & \textbf{AS} \\
\midrule
\textbf{SAU}                & 0.839   & 0.154     & 5.656              \\
\textbf{SAU\&LAU}            & 0.867   & 0.125     & 5.801              \\
\textbf{SAU\&EU (avgR)}     & 0.875   & 0.117     & 5.724              \\
\textbf{SAU\&EU (maxR)}      & 0.885   & 0.107     & 5.754              \\
\textbf{SAU\&LAU\&EU (avgR)} & \textbf{0.914}   & \textbf{0.078}     & 5.816              \\
\textbf{SAU\&LAU\&EU (maxR)}  & 0.909   & 0.083     & \textbf{5.879}               \\ 
\bottomrule
\end{tabular}
\label{Tab6}
\end{table}


\begin{table*}[]
\caption{Evaluation of Planning under Different Perceptual Limitations.}
\centering
\begin{tabular}{m{2cm}<{\centering} | m{0.8cm}<{\centering}  m{0.8cm}<{\centering}  m{0.8cm}<{\centering} | m{0.8cm}<{\centering}  m{0.8cm}<{\centering}  m{0.8cm}<{\centering} | m{0.8cm}<{\centering}  m{0.8cm}<{\centering}  m{0.8cm}<{\centering} | m{0.8cm}<{\centering}  m{0.8cm}<{\centering}  m{0.8cm}<{\centering}}
\toprule
                    & \multicolumn{3}{c|}{\textbf{occlusion}}  & \multicolumn{3}{c|}{\textbf{noise($\sigma$=0.01)}} & \multicolumn{3}{c|}{\textbf{noise($\sigma$=0.1)}} & \multicolumn{3}{c}{\textbf{occlusion\&noise($\sigma$=0.2)}} \\ \cmidrule(l){2-13} 
                    & \textbf{SR} & \textbf{CR} & \textbf{AS} & \textbf{SR}         & \textbf{CR}         & \textbf{AS}         & \textbf{SR}         & \textbf{CR}         & \textbf{AS}        & \textbf{SR}             & \textbf{CR}            & \textbf{AS}            \\
\midrule
\textbf{non-UAP}        & 0.786       & 0.206       & 5.559       & 0.802               & 0.190               & 5.570               & 0.797               & 0.195               & 5.598              & 0.766                   & 0.227                  & 5.704                  \\
\textbf{SAU}        & 0.833       & 0.159       & 5.657       & 0.846               & 0.146               & 5.672               & 0.823               & 0.169               & 5.707              & 0.797                   & 0.195                  & 5.835                  \\
\textbf{SAU\&LAU}    & 0.857       & 0.135       & 5.807       & 0.867               & 0.125               & 5.807               & 0.844               & 0.148               & 5.831              & 0.812                   & 0.180                  & 6.011                  \\
\textbf{SAU\&EU}     & 0.880       & 0.112       & 5.756       & 0.867               & 0.125               & 5.757               & 0.893               & 0.099               & 5.765              & 0.865                   & 0.128                  & 5.847                  \\
\textbf{SAU\&LAU\&EU} & 0.911       & 0.081       & 5.839       & 0.904               & 0.089               & 5.819               & 0.883               & 0.109               & 5.859              & 0.859                   & 0.133                  & 5.974                 \\
\bottomrule
\end{tabular}
\label{Tab7}
\end{table*}

\subsubsection{Analysis of Comprehensive Risk Model}
Table \ref{Tab6} provides a comprehensive comparison of different uncertainty modeling approaches for UAP. For SAU\&LAU, the UAP design employs the (maxR) strategy with superior performance, as shown in Table \ref{Tab2}. Similar to SAU\&EU, the results of (avgR)(\textit{mc.}) and (maxR)(\textit{mc.}) strategies, which perform well in Table \ref{Tab3}, are shown.
For SAU\&LAU\& EU, the same strategies utilized in SAU\&LAU and SAU\&EU are respectively employed to implement \(g_{ale}()\) and \(g_{epi}()\) in Eq. (\ref{eq16}).

The UAP approach, considering SAU, LAU, and EU, yields the best performance, with an 8.9\% increase in SR compared to solely modeling SAU. This underscores the benefits of modeling all uncertainties, leveraging LAU to capture diverse distributions of traffic participant motion patterns, and addressing the reliability issues of individual prediction models using EU modeling.

\subsection{Testing under Limited Perception}
This section assesses the impact of limited perception on UAP performance, highlighting the system's robustness in scenarios with substantial distributional shifts.

To simulate perceptual limitations, common challenges were introduced to the UAP process. Occlusion, which stems from the limited perception range of vehicles, obscures some traffic participants. UAP performance was investigated under occlusion from road boundaries or obstacles. Additionally, perceptual inaccuracies in capturing motion states of traffic participants were simulated by introducing Gaussian noise (\(\sigma=0.01\) and \(\sigma=0.1\)) into the prediction and planning processes. An extreme test case combined occlusion and higher noise (\(\sigma=0.2\)) to evaluate performance under severe distributional shifts.

Table \ref{Tab7} shows that UAP models that incorporate multiple uncertainties consistently surpass those that only account for SAU or disregard uncertainties, particularly under heightened perceptual constraints. The incorporation of EU notably enhances planning performance because of its effectiveness in estimating model inadequacies, and addresses the unreliability stemming from unfamiliar or anomalous inputs. This underscores the pivotal role of EU in augmenting the adaptability of the prediction-planning system in complex scenarios.

\subsection{Analysis of Typical Cases}

\begin{figure*}
    \color{black}
    \centering
    \subfloat[\fontsize{7}{7}\selectfont]{\includegraphics[width=4.5cm]{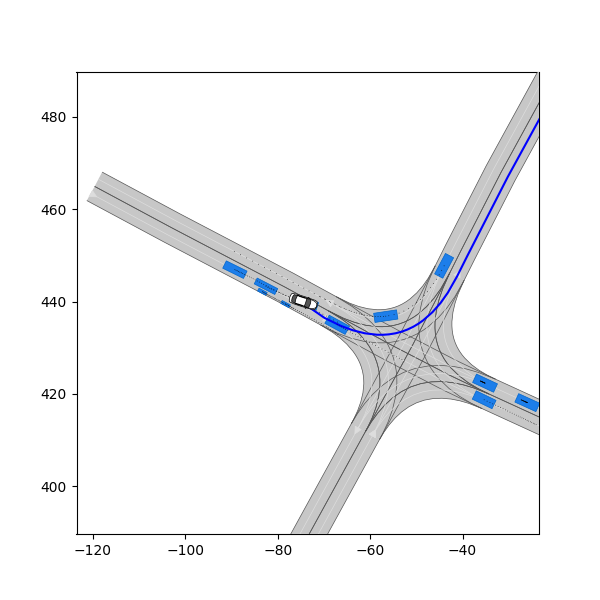}}
    \hfill
    \subfloat[\fontsize{7}{7}\selectfont]{\includegraphics[width=4.5cm]{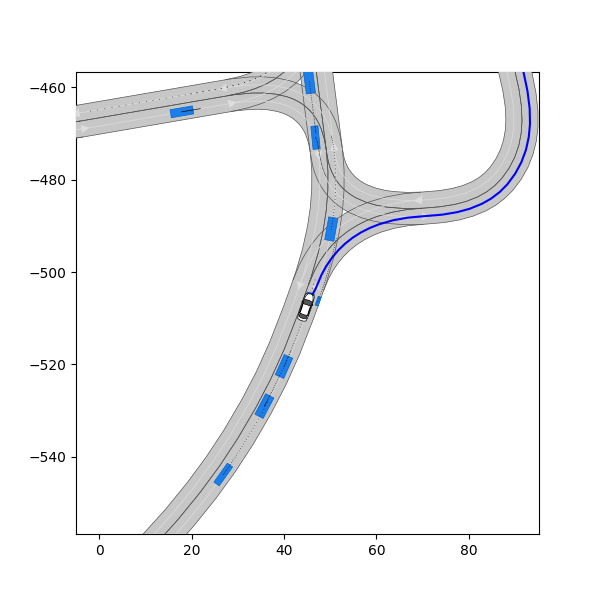}}
    \hfill
    \subfloat[\fontsize{7}{7}\selectfont]{\includegraphics[width=4.5cm]{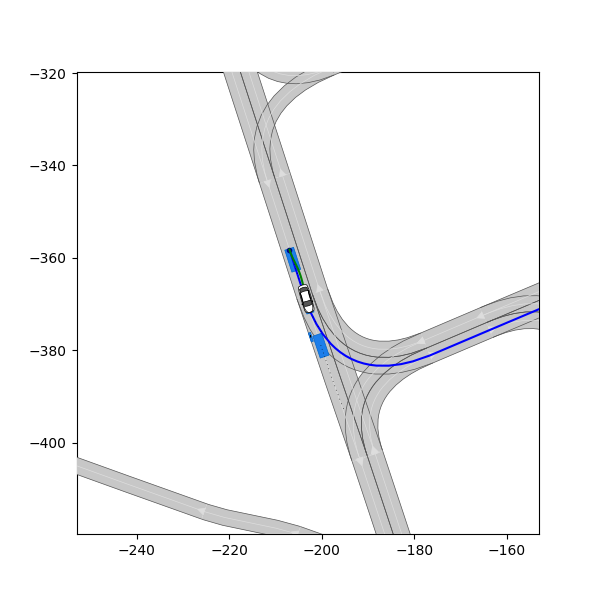}}
    \hfill
    \subfloat[\fontsize{7}{7}\selectfont]{\includegraphics[width=4.5cm]{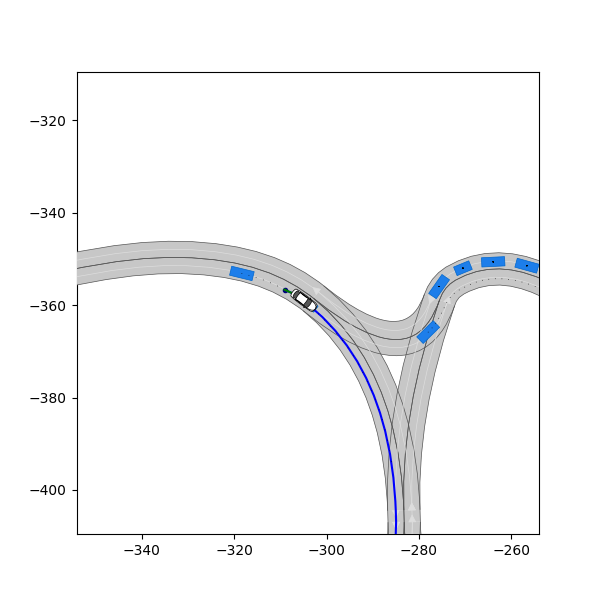}}
    \caption{Typical scenarios. The ego vehicle is represented as a white car, whereas other traffic participants are depicted with blue rectangular.}
    \label{Figr1}
    \color{black}
\end{figure*}

\begin{figure*}
    \color{black}
    \centering
    \subfloat[\fontsize{7}{7}\selectfont]{\includegraphics[width=4.5cm]{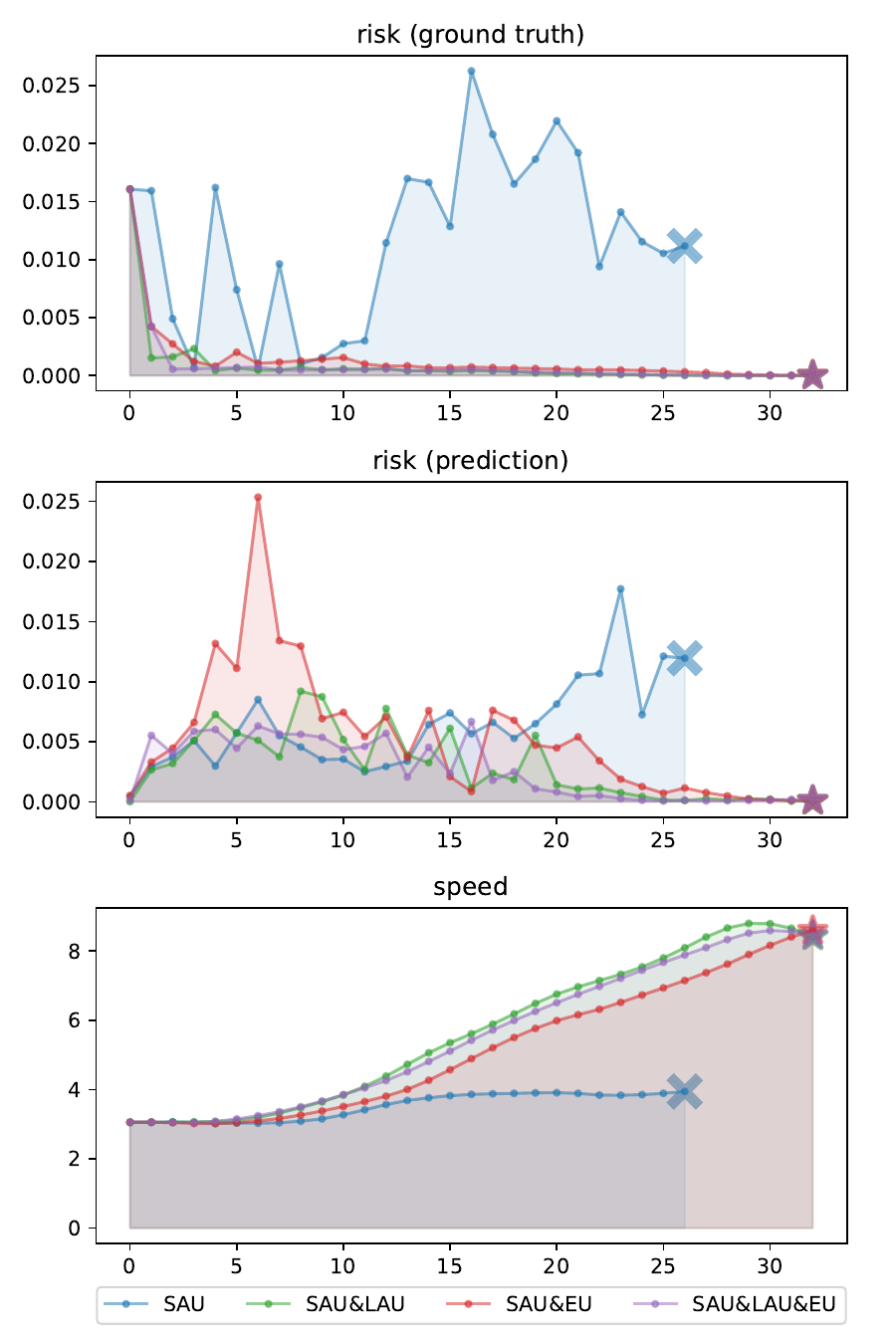}}
    \hfill
    \subfloat[\fontsize{7}{7}\selectfont]{\includegraphics[width=4.5cm]{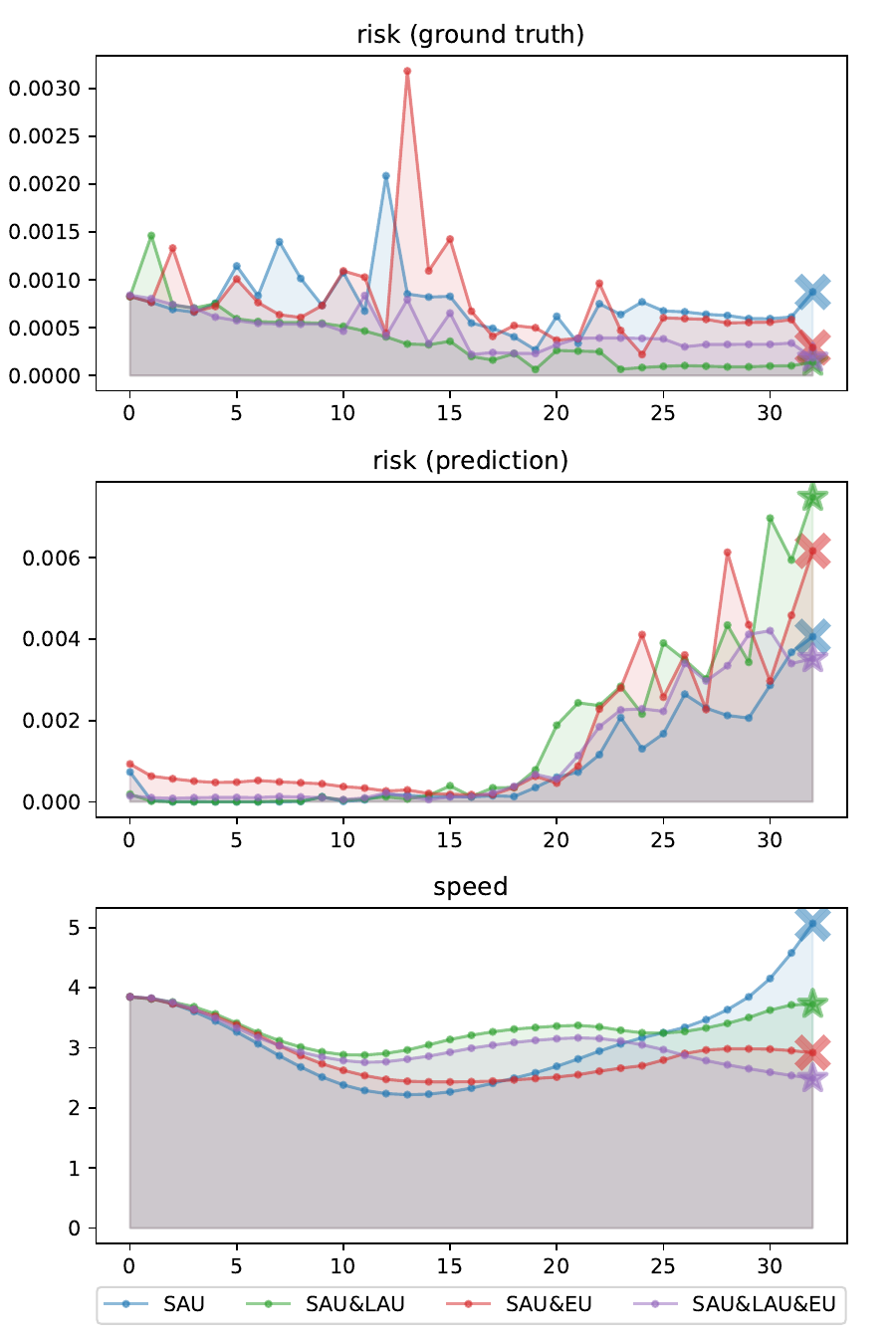}}
    \hfill
    \subfloat[\fontsize{7}{7}\selectfont]{\includegraphics[width=4.5cm]{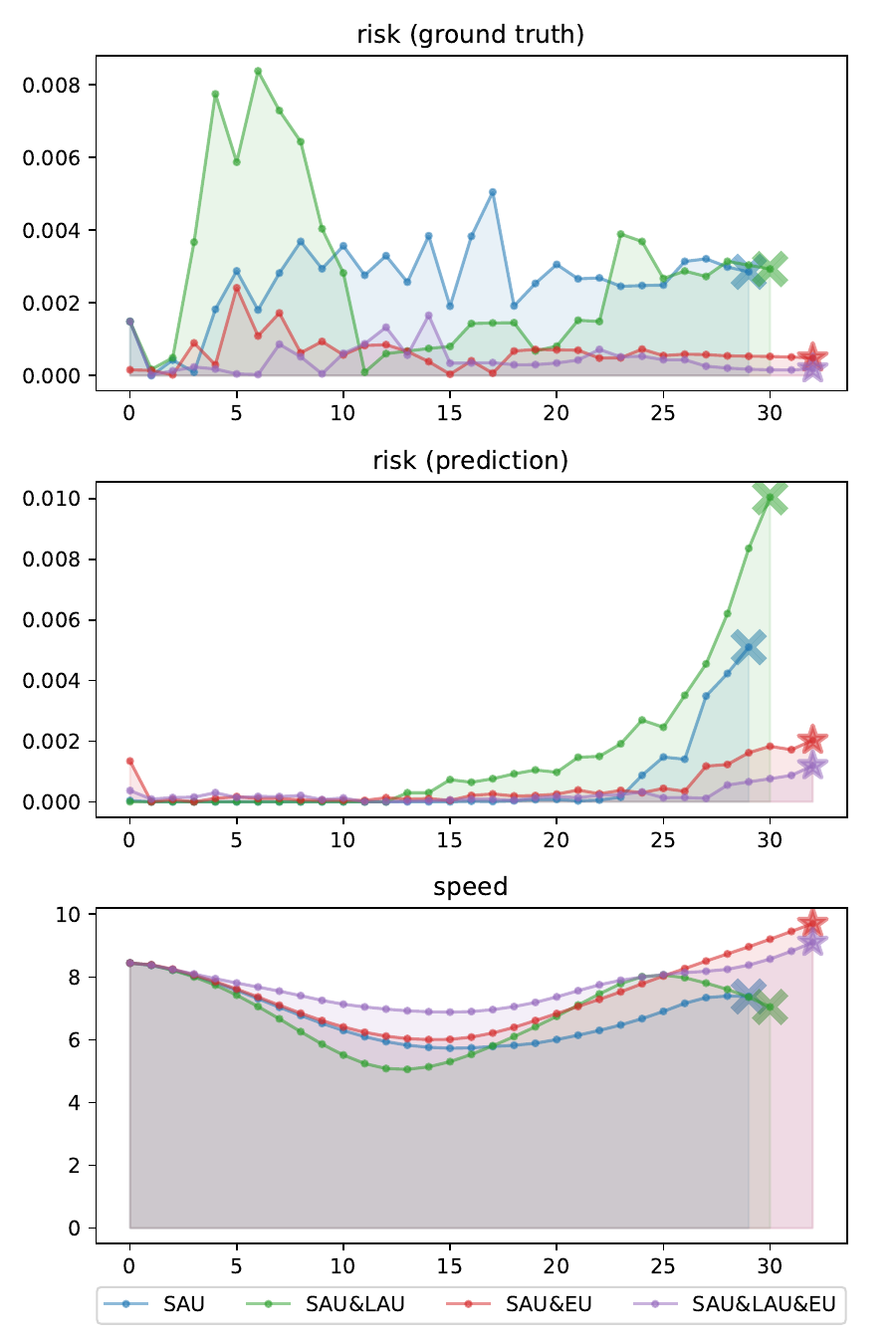}}
    \hfill
    \subfloat[\fontsize{7}{7}\selectfont]{\includegraphics[width=4.5cm]{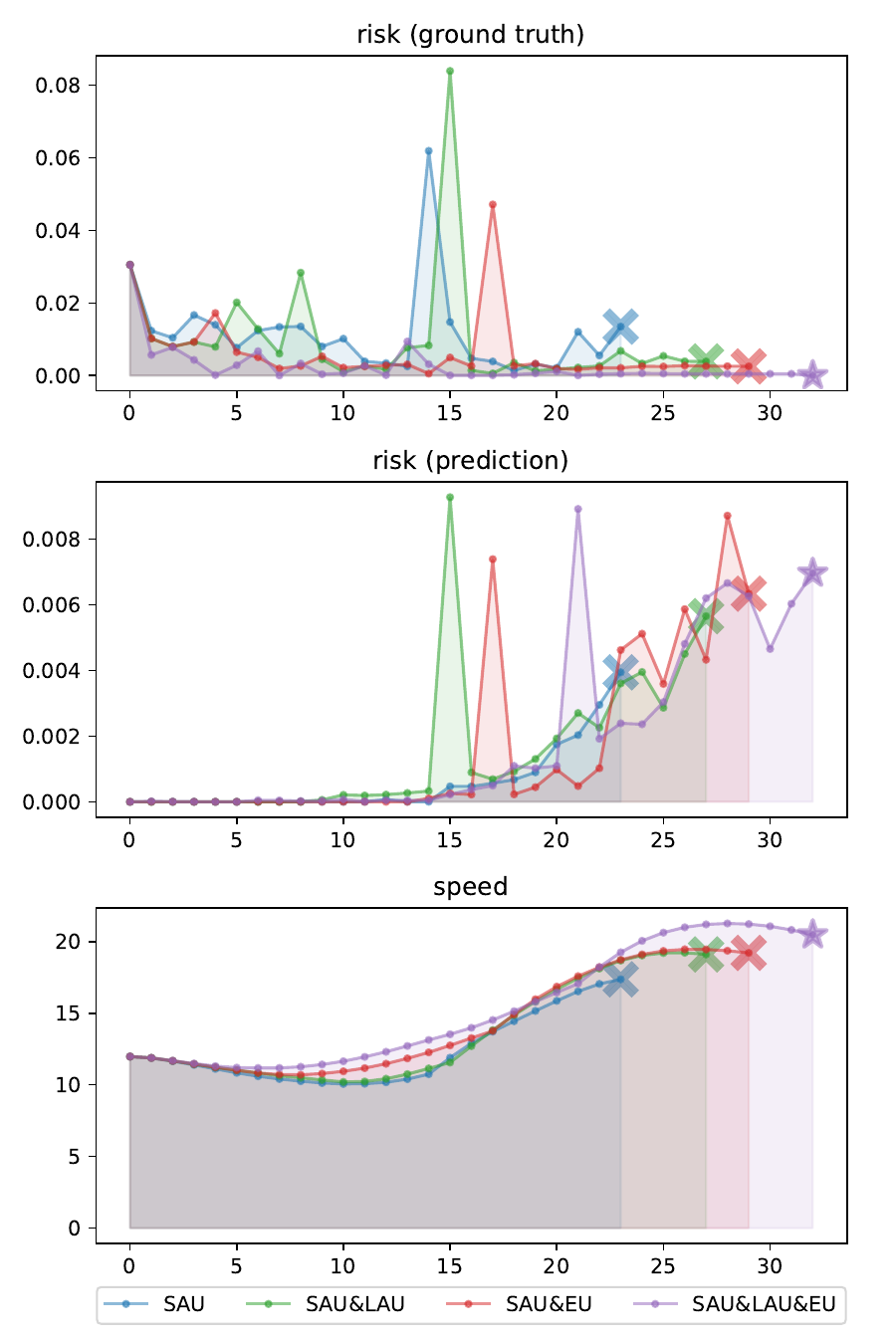}}
    \captionsetup[subfloat]{font=tiny}

    \caption{
    Results from various UAP methods for the four scenarios shown in Fig. \ref{Figr1}. At the endpoint of each curve, \ding{72} indicates successful passage, whereas \(\times\) indicates a collision. The x-axis represents time, and the y-axis represents the focused metrics:
    The first row depicts the risk (ground truth) incurred during the ego vehicle's travel. The second row records the variation of risk (prediction). Additionally, speeds are shown in the third row.
    }
    \label{Figr2}
    \color{black}
\end{figure*}

This subsection provides an in-depth analysis of typical cases to better understand the performance of different UAP methods.
Fig. \ref{Figr1} shows results for complex interactions, such as intersections and T-junctions. The ego vehicle is expected to navigate while avoiding collisions with other traffic participants or road boundaries. Figure \ref{Figr2} shows the test results for UAPs under different uncertainty conditions. The concept of 'risk (ground truth)' is used for an intuitive understanding of planning safety, estimated through the distribution $P_{gt_{i}}^{t} \sim \mathcal{N}(\mu_{i}^{t}, \Sigma_{i}^{t})$, where $\mu_{i}^{t}$ denotes the real trajectory of $A_{i}$, and $\Sigma_{i}^{t}$ is set as a constant for modeling risk.
Effective predictors are considered to aid the ego vehicle in navigating test scenarios while maintaining a low risk (ground truth). The 'risk (prediction)' reflects the vehicle's perceived risk, based on its prediction results and uncertainty estimates. The following is a case-by-case analysis:

Case (a) features a left-turn scenario at an intersection. In Fig. \ref{Figr2}-(a), the UAP considering only SAU fails, showing unreliable initial risk estimation, leading to slower operation and a collision. In contrast, UAPs integrating LAU or EU successfully navigate this scenario, highlighting the benefits of multi-uncertainty modeling.

Case (b) involves a right turn at an intersection (Fig. \ref{Figr2}-(b)). Modeling LAU in prediction and planning reduces actual risks and ensures successful scenario navigation.

Case (c) deals with a left turn at a T-junction, with close vehicles both in front and behind. Introducing EU modeling and risk modeling is advantageous. A single prediction model shows unreliable performance, resulting in an elevated risk level and collision, whereas deep ensemble enhances risk awareness, leading to safe and efficient navigation.

Case (d) shows the benefits of considering all three types of uncertainty. UAPs that do not collectively model SAU, LAU, and EU experience fluctuating risk levels and collisions. Conversely, the UAP based on SAU\&LAU\&EU avoids accidents effectively, particularly evident at critical moments, such as at 2.1 s, where high predicted risk leads to swift and safe maneuvering away from collisions with the following vehicle.

\section{Conclusion}
This study provides a comprehensive perspective and comparison of uncertainty-aware prediction  and planning in autonomous driving systems. We offer a systematic analysis of uncertainties in prediction, encompassing SAU, LAU, and EU, and investigate their sources of uncertainty and modeling approaches. Based on rational modeling, a unified framework for prediction and planning was introduced. This framework facilitates risk modeling for different uncertainties and their combinations, thereby enabling efficient planning under complex traffic conditions. This allows for concurrent modeling and consideration of the SAU, LAU, and EU.
Utilizing this framework, we conducted a thorough comparison of diverse uncertainty and risk-modeling methods, encompassing many representative existing methods. The experimental findings underscore the advantages of integrating uncertainty into planning and reveal improvements in system performance when multiple sources of uncertainty are considered, in contrast to single-source models. Moreover, the proposed framework facilitates the development of methods for UAP that surpass existing uncertainty-aware risk models. In conclusion, the introduction of LAU and EU modeling has emerged as impactful, improving the accuracy and reliability of UAP in diverse traffic scenarios.
Future studies will focus on incorporating advanced autonomous driving technologies and paradigms to refine uncertainty modeling, aiming to improve decision-making in complex driving scenarios.


\bibliographystyle{IEEEtran}
\bibliography{ref}
\vspace{-1cm}

\end{document}